\documentclass{bmvc2k}

\title{Gaussian Splatting in Mirrors: Reflection-Aware Rendering via Virtual Camera Optimization}

\addauthor{Zihan Wang}{zihan.1.wang@aalto.fi}{1}
\addauthor{Shuzhe Wang}{shuzhe.wang@aalto.fi}{1}
\addauthor{Matias Turkulainen}{matias.turkulainen@aalto.fi}{1,3}
\addauthor{Junyuan Fang}{junyuan.fang@aalto.fi}{1}
\addauthor{Juho Kannala}{juho.kannala@aalto.fi}{1,2}

\addinstitution{
Aalto University
}
\addinstitution{
University of Oulu
}

\addinstitution{
ETH Zurich
}
\runninghead{Wang ET AL.}{Gaussian Splatting in Mirrors}

\usepackage{multirow}
\usepackage[normalem]{ulem}
\usepackage{amsfonts}

\usepackage{times}
\usepackage{epsfig}
\usepackage{amsmath}
\usepackage{amssymb}
\usepackage{booktabs}
\usepackage{multirow}
\usepackage{gensymb}
\usepackage{bbding}
\usepackage{soul}
\usepackage{multirow}
\usepackage{array}
\usepackage{enumitem}
\usepackage{pifont}
\usepackage{graphicx}
\usepackage{booktabs}
\usepackage{wrapfig}
\begin{document}

\maketitle

\begin{abstract}
Recent advancements in 3D Gaussian Splatting (3D-GS) have revolutionized novel view synthesis, facilitating real-time, high-quality image rendering. However, in scenarios involving reflective surfaces, particularly mirrors, 3D-GS often misinterprets reflections as virtual spaces, resulting in blurred and inconsistent multi-view rendering within mirrors. Our paper presents a novel method aimed at obtaining high-quality multi-view consistent reflection rendering by modelling reflections as physically-based virtual cameras. We estimate mirror planes with depth and normal estimates from 3D-GS and define virtual cameras that are placed symmetrically about the mirror plane. These virtual cameras are then used to explain mirror reflections in the scene. To address imperfections in mirror plane estimates, we propose a straightforward yet effective virtual camera optimization method to enhance reflection quality. We collect a new mirror dataset including three real-world scenarios for more diverse evaluation. Experimental validation on both Mirror-Nerf and our real-world dataset demonstrate the efficacy of our approach. We achieve comparable or superior results while significantly reducing training time compared to previous state-of-the-art. We release our code as open-source at: \url{https://github.com/rzhevcherkasy/BMVC24-GSIM}.

\end{abstract}

%-------------------------------------------------------------------------
\section{Introduction}
\label{sec:intro}

3D Gaussian Splatting (3D-GS)~\cite{kerbl20233d} has recently made significant advancements in the field of novel view synthesis (NVS)~\cite{lu2023scaffold,yan2023multi,luiten2023dynamic,zheng2023gps,zhu2023fsgs,wu20234d} and scene reconstruction~\cite{li2024gaussianbody,yang2023deformable,dust3r_cvpr24,lin2024vastgaussian,lyu20243dgsr,charatan2023pixelsplat}. The method employs an explicit Gaussian based scene representation and novel differentiable rasterization algorithm, enabling high fidelity rendering quality that rivals that of Neural Radiance Field (NeRF)~\cite{mildenhall2021nerf} based methods, while significantly reducing rendering times. Although 3D-GS performs well in NVS, it encounters difficulties in handling specular reflections, particularly when mirrors or other reflective objects are present in the scene. The method tends to misinterpret reflections as virtual spaces behind mirrors, leading to multi-view inconsistencies. This results in blurry and disordered renderings of mirrors and their edges, which compromises the overall quality of the novel view synthesis.

Prior work based on NeRF has explored reflection-aware rendering to tackle specular reflection. Ref-NeRF~\cite{verbin2022ref} substitutes the origninal NeRF's ray-marching parametrization of view-dependent radiance with a representation that includes reflected radiance, aiming to directly account for reflective surfaces during optimization. MS-NeRF~\cite{yin2023multi} represents mirror reflections as a group of feature fields in a parallel sub-space, enhancing the model's ability to handle complex reflective scenarios. NeRFReN~\cite{guo2022nerfren} instead models both transmitted and reflective components in a scene by two separate radiance fields, providing a more nuanced depiction of light interactions. Among these advancements, Mirror-NeRF~\cite{zeng2023mirror} stands out as the closest state-of-the-art method. It estimates the direction of reflected rays in the scene and employs Whitted Ray Tracing~\cite{whitted2005improved} for a more physically accurate ray-tracing model for reflective scenes. Despite these innovations, NeRF-based approaches are severely limited by their long training times and inability to achieve real-time rendering speeds on common devices.

In 3D-GS literature, the issue of reflections has not been extensively addressed. Concurrent work Mirror-3DGS~\cite{meng2024mirror}, adopts a virtual camera to render mirror regions, but Mirror-3DGS requires additional depth supervision tailored specifically for mirrors. Another concurrent work, MirrorGaussian~\cite{liu2024mirrorgaussian}, leverages dual-rendering strategy to render both real-world Gaussians and the virtual ones from the mirror space. In contrast, our method solves the reflection problem using virtual camera rendering. Furthermore, current research into mirror reflections encounters significant challenges due to a lack of diverse real-world scene data. Although Mirror-NeRF includes three real-world indoor scenes featuring mirrors, the dataset is limited as the mirrors have the same size and shape, which does not fully represent the complexity of diverse real-world environments. This limitation highlights the need for more varied datasets to better understand and address the challenges in reflection rendering in real-world settings.

To address the issue of mirror reflections in 3D Gaussian Splatting (3D-GS), we develop a progressive training pipeline for mirror rendering that utilizes virtual camera rendering inspired by~\cite{rodrigues2010camera}. We first predict a mirror plane equation in world coordinates through depth and normal estimations in 3D-GS, which enables us to further derive the virtual camera pose. A rendered image of the 3D-GS scene containing a mirror in view is then created as a fusion of a reflected image obtained from the virtual camera render and a non-reflected image which is obtained by traditional 3D-GS rasterization into the current camera view. To enhance the quality of the virtual camera rendering, the predicted mirror plane is refined by virtual camera pose optimization during optimization. To address the problem of insufficient real-world scene data, we collected a new dataset containing three real-world scenes with mirrors. Our dataset encompasses scenes of varying scales, incorporating mirrors of different sizes and shapes. We believe this new dataset can progressively provide training and testing environments ranging from simple to complex, thereby better facilitating the evaluation of method effectiveness.

Our contributions can be summarized as follows:
\begin{itemize}
    \item We propose a rendering method based on 3D-GS for scenes containing mirror reflections. Mirror reflections are explained by renders onto virtual cameras placed around a mirror plane. Output images of camera views containing mirrors are a fusion of virtual camera renders and traditional 3D-GS renders. Our method achieves high-quality mirror reflection rendering and maintains the real-time rendering capabilities of 3D-GS with minimal added memory or computation needs.
    \item We present a dataset featuring real-world scenes with mirrors, encompassing various scales and mirror sizes. Unlike existing datasets, ours progressively includes mirror scenes of increasing complexity, offering challenging scenes for evaluation of the method's effectiveness.
    \item Experimental results on both synthetic and real-world scenes demonstrate that our method matches or even exceeds state-of-the-art techniques across multiple metrics while preserving the high rendering quality of 3D-GS.
\end{itemize}

\begin{figure*}
\begin{center}
\includegraphics[width=0.95\linewidth]{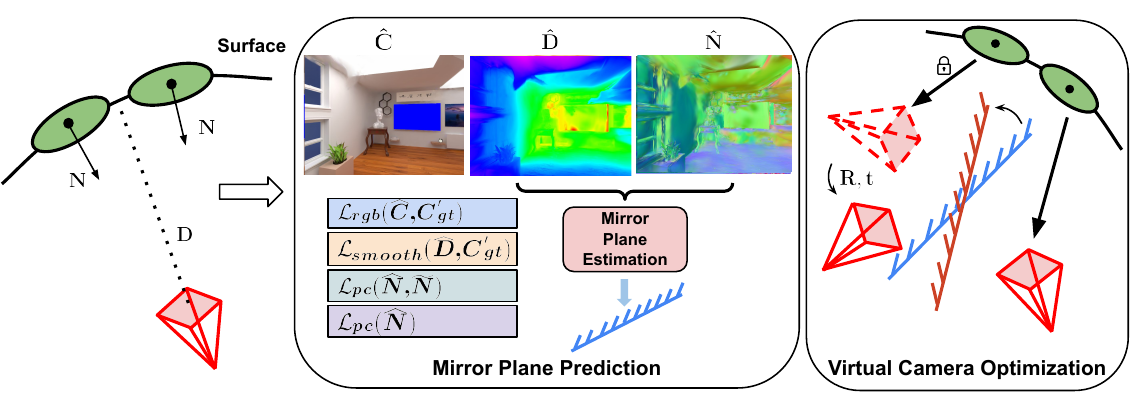}
\end{center}
    \caption{Pipeline Overview: We extend 3D-GS with depth and normal supervision to initialize a mirror plane in Sec. \ref{sec:Mirror Plane Prediction}. Blue region in $\hat{C}$ indicates the mirror. We render both real and virtual camera viewpoints and combine them into a single image in Sec \ref{sec:Virtual Camera Rendering}. We further refine virtual camera positions during optimization to achieve photo-realistic mirror reflections in Sec. \ref{sec:Virtual Camera Optimization}.}
\label{fig:overview}
\end{figure*}

\section{Method}\label{sec:method}

This section details the method of the proposed pipeline. We first review the 3D Gaussian Splatting~\cite{kerbl20233d}  in Sec.~\ref{sec: gaussian}. Sec~\ref{sec:Mirror Plane Prediction} introduces the mirror plane calculation from the estimated depth and normal maps. Sec~\ref{sec:Virtual Camera Rendering} defines the virtual camera pose by the estimated mirror plane equation and Sec~\ref{sec:Virtual Camera Optimization} refine the mirror plane via the virtual camera optimization. The training objective is presented in Sec~\ref{sec:Progressive Training}. The main pipeline is illuminated in ~Fig. \ref{fig:overview}.
% The key to obtaining high-quality reflection rendering lies in augmenting the vanilla 3DGS renderer to correctly render mirror regions seen in the various training camera views. We assume that for each training view, we have a segmentation mask obtained from an off-the-shelf predictor~\cite{kirillov2023segment} to give location of a mirror. Our approach extends the 3DGS \cite{kerbl20233d} renderer to render normal and depth maps per camera view and we regularize the depth and normal predictions at segmented mirror locations as described in Sec. \ref{sec:Mirror Plane Prediction}. We then use the depth and normal predictions at mirror locations to intialize a mirror plane about which virtual cameras are positioned and use these virtual cameras to synthesize mirror reflection regions in Sec. \ref{sec:Virtual Camera Rendering}. We further optimize the virtual camera positions obtained from depth and normal estimations using camera optimization in Sec. \ref{sec:Virtual Camera Optimization} to achieve more robust renders.

\subsection{3D Gaussian Splatting}\label{sec: gaussian}
We build our method on 3D Gaussian Splatting \cite{kerbl20233d} which represents a scene by a collection of differentiable Gaussian distributions $\mathcal G_i = \{ (\mu_i, \Sigma_i, O_i, \theta_i ) \}_i$, 
% \begin{equation*}
%   \mathcal G_i = \{ (\mu_i, \Sigma_i, o_i, \theta_i ) \}_i
% \end{equation*}
with means and covariance matrices
 $(\mu_i, \Sigma_i)$, opacities $O_i$, and view-dependent colours $C_i \in \mathcal C^{N_{\rm sh}}$ represented via spherical harmonic coefficients $N_{\rm sh}$. To render individual views from the Gaussian scene representation, Gaussians are projected into screen space as 2D Gaussians using the current camera view matrix, sorted by z-depth, and alpha-blended to produce pixel colors $\hat{C}$:
 \begin{align}
  {\hat{C}}  & = \sum_{i \in N} {{c}}_{i}\alpha_i T_i, \textrm{ where } T_i = \prod_{j = 1}^{i - 1} (1- \alpha_j),
\end{align}
\noindent where $T_i$ is the accumulated transmittance at the rendered pixel $p$ and $\alpha_{i}$ consists of the $i^{th}$ Gaussian's blending term located in view space with position $\hat{\mu_i}$:

\begin{align}
  {\alpha_i}  & = O_i \cdot \exp{\left(\frac{1}{2}({p}-{\hat{\mu}}_i){\Sigma}_i^{-1}({p}-{\hat{\mu}}_i)\right)}.
\end{align}
The process of projection and rendering is parallelized allowing for real-time rendering. 

\subsection{Virtual Camera Definition}\label{sec:Virtual Camera Rendering}

Prior work \cite{zeng2023mirror} address mirror reflections in the NeRF context using Whitted-Style Ray Tracing. However, performing ray tracing on a 3D-GS scene is computationally expensive, necessitating the exploration of alternative methods. Inspired by ~\cite{rodrigues2010camera}, we observe that a region of an image from a real camera containing a reflection from a mirror can be explained by an alternative image obtained by a virtual camera placed symmetrically about a mirror plane. In fact, with perfect mirror reflections, the image seen of a reflection from the real camera and the image seen from that of the virtual camera are identical. The virtual camera is symmetrically posed behind the mirror as shown in Fig. \textcolor{red}{2} and views the same Gaussian scene.

\begin{wrapfigure}{r}{0.5\textwidth}
\begin{center}
    \hspace{-15pt}\includegraphics[width=1.05\linewidth]{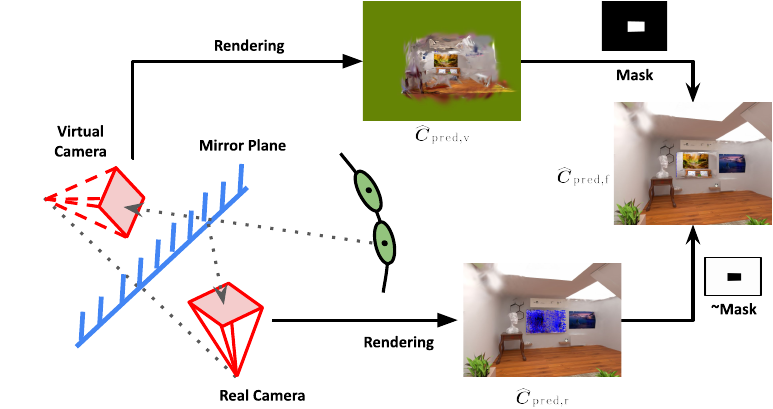}
    \vspace{10pt}
   \caption{The process of virtual camera rendering in Sec. \ref{sec:Virtual Camera Rendering}. 
   }
\end{center}
\vspace{-30pt}
\label{fig:virtual_camera}
\end{wrapfigure}

Given a real camera viewpoint with the extrinsic matrix $ \left [ \bf{R | t} \right ] \in \mathbb{R}^{3 \times 4}$ and the mirror plane $[\mathbf{n}, o]$ (discussed in Sec.~\ref{sec:Mirror Plane Estimation}), we obtain the extrinsic matrix of virtual camera $ \left [ \bf{R}'| \bf{t}' \right ] \in \mathbb{R}^{3 \times 4}$ following ~\cite{rodrigues2010camera}:

\begin{equation*}\label{eq: virtual transform}
\begin{bmatrix}
 \bf{R}' & \bf{t}' \\
 0 & 1
\end{bmatrix}  =\begin{bmatrix}
\mathbf{I}- 2\mathbf{n}\mathbf{n}^{T}   & 2o\mathbf{n} \\
 0 &1
\end{bmatrix} \times \begin{bmatrix}
 \bf{R} & \bf{t}\\
0  &1 
\end{bmatrix}.
\end{equation*}

Once the virtual camera pose is determined, we render the mirror-reflection image using this virtual camera and standard 3D-GS forward rasterization. Non-reflective regions are rendered by the corresponding real camera. We fuse both regions using segmented mirror masks~\cite{kirillov2023segment} to generate the output image $\hat{C}_{\text{pred,f}}$:
\begin{equation}\label{eq: fuse}
\hat{C}_{\text{pred,f}}(k) = \begin{cases} 
\hat{C}_{\text{pred,v}}(k), & \text{if } k \in \mathcal{P}_{cam}\cap \mathcal{P}_{M} \\
\hat{C}_{\text{pred,r}}(k), & \text{if } k \in \mathcal{P}_{cam}\cap \overline{\mathcal{P}_{M}}
\end{cases},
\end{equation}

where \( \hat{C}_{\text{pred,r}} \) is the rendered region obtained from the real camera, \( \hat{C}_{\text{pred,v}} \) is the rendered with the virtual camera. \( \mathcal{P}_{\text{cam}} \) denotes all the pixels in the rendered image, and \( \mathcal{P}_{M} \) is the set of pixels belonging to the masked mirror region.

\subsection{Mirror Plane Prediction}\label{sec:Mirror Plane Prediction}
To obtain the mirror plane in world coordinates, we extend the original 3D-GS rendering with with depth and normal rasterization. Following DN-Splatter \cite{turkulainen2024dnsplatter}, we render depth and normal maps from the 3D-GS scene for each camera view.

\noindent\textcolor{black}{\textbf{Color optimization.}} During this stage, we only render RGB image $\hat{C}_{\text{pred,r}}$ from real camera. In order to reconstruct the mirror surface, we replace the mirror region color on gt image with a constant color to obtain $C'_{gt}$. We use $\mathcal{L}_{rgb}(\hat{C}_{\text{pred,r}},C'_{gt})$ to optimize the color where $\mathcal{L}_{rgb}$ is the regularization loss  proposed in 3D-GS~\cite{kerbl20233d}.

\noindent\textcolor{black}{\textbf{Depth estimation and regularization.}}
Given a camera view with an extrinsic matrix $ \left [ \bf{R} | \bf{t} \right ] \in \mathbb{R}^{3 \times 4}$ and intrinsics $\mathbf{K} \in \mathbb{R}^{3 \times 3}$, we estimate depth maps from the center of each 3D Gaussians \(\mathbf{\mu}_i \) with the following equation:
\begin{equation}\label{eq:} \mathbf{\mu}'_i= \left[ x_i,y_i,d_i \right]^T = \mathcal{T}(\mathbf{\mu}_i, \mathbf{R}, \mathbf{t}, \mathbf{K}),
\end{equation}
where $\mathcal{T} (\cdot)$ is the projection equation. We then use the z-depth $d_i$ in $\bf{\mu}'_i$ and render per-pixel depths \( \hat{D} \) using alpha-compositing:
\begin{equation}\label{eq: depth render}
\hat{D} = \sum_{i \in N }^{} d_i \alpha_i\prod_{j=1}^{i-1}(1-\alpha_j)   ,
\end{equation}
where $\alpha_i$ is the blending coefficient for the \( i \)-th Gaussian. 
% We mainly focus on indoor datasets with planar sufaces, therefore, depth estimates at nearby pixels should be similar. We adopt a smoothness prior on depth estimates following ~\cite{guo2022nerfren,turkulainen2024dnsplatter} using an RGB-weighted depth smoothness regularization loss which penalizes non-similar depth predictions at textureless regions:
We primarily concentrate on indoor scenes, which typically feature numerous planar surfaces. Therefore, depth estimates for adjacent pixels are expected to be similar. To enforce this consistency, we employ a smoothness prior to depth estimation~\cite{guo2022nerfren,turkulainen2024dnsplatter}. Specifically, we use an RGB-weighted depth smoothness regularization loss that penalizes dissimilar predictions in textureless areas:
\begin{equation}\label{eq: depth smooth}
\mathcal{L}_{smooth} = \sum_{k \in \mathcal{P}_{cam}}^{} \sum_{q \in \mathcal{N}(k)}^{} exp(-\gamma \left |C'_{gt}(k)-C'_{gt}(q) \right |) \left | \hat{D}_{pred,r}(k)-\hat{D}_{pred,r}(q) \right |,
\end{equation}
where \( k \) and \( q \) represent pixels, \( \mathcal{N}(k) \) represents the 4 horizontal/vertical neighbour pixels around \( k \), and \( \gamma \) is a hyper-parameter.

\noindent\textcolor{black}{\textbf{Normal estimation and regularization.}}
3D Gaussians do not inherently contain a normal direction. However, previous methods ~\cite{jiang2023gaussianshader, turkulainen2024dnsplatter} have observed that during 3D-GS optimization process, Gaussians gradually become flatter and approach a planar shape. Therefore, the shortest scaling axis of a Gaussian's covariance matrix can be considered as a good estimate of a normal direction. Similar to depth rendering, we obtain per-pixel normal maps following:
\begin{equation}\label{eq: normal render}
\hat{N} = \sum_{i \in N }^{} \mathbf{n}_i \alpha_i\prod_{j=1}^{i-1}(1-\alpha_j)  ,
\end{equation}
where \( \mathbf{n}_i \in \mathbb{R}^{3} \) is the normalized vector representing the direction of the shortest axis of the \( i \)-th Gaussian. We leverage the above depth estimates for supervision to ensure consistent normal estimates at nearby pixels. We encourage consistency between the rendered normal \( \hat{N} \) and the pseudo-normal \( \tilde{N} \), computed from the gradient of rendered depths \( \hat{D} \) under the local planarity assumption. The normal consistency is quantified as an L1 loss:
\begin{equation}\label{eq: loss normal gradient}
\mathcal{L}_{n} = \left \| \hat{\mathbf{N}} - \tilde{\mathbf{N}} \right \|  ,
\end{equation}
where $\tilde{\mathbf{N}}$ is the pseudo ground truth normal estimate derived from the gradient of the depth map~\cite{huang20242d}:

% \begin{equation}\label{eq: pseudo-normal}
% \tilde{N}(i) = \frac{1}{\sqrt{1 + \left(\frac{\partial d_i}{\partial x_i}\right)^2 + \left(\frac{\partial d_i}{\partial y_i}\right)^2}} \begin{bmatrix} -\frac{\partial d_i}{\partial x_i} \\ -\frac{\partial d_i}{\partial y_i} \\ 1 \end{bmatrix}
% \end{equation}

\begin{equation}\label{eq: pseudo-normal}
\tilde{\mathbf{N}}(x,y) =\frac{\bigtriangledown_xd \times  \bigtriangledown_yd }{|\bigtriangledown_xd \times  \bigtriangledown_yd|} ,
\end{equation}
and $d$ are nearby depth points of pixel $(x,y)$.
% TODO: ADD EQUAITON FOR THE PSEUDO-GROUND-TRUTH-NORMAL-FROM-DEPTH-GRADIENT
% Done

In addition to the regularization with pseudo ground truth normals, we enforce that segmented mirror regions to have the same normal direction with a planar assumption. Specifically, during training we randomly sample \( p \) pixels from the segmented mirror region and penalize differences in angular normal estimates with the planar-constraint loss:

\begin{equation}\label{eq: normal consistent}
\mathcal{L}_{pc} = \frac{1}{\mathbf{N}_p^2} \sum_{k=1}^{\mathbf{N}_p} \sum_{q=1}^{\mathbf{N}_p} (1-\cos(\langle \mathbf{n_k}, \mathbf{n_q} \rangle)),
\end{equation}
where $\mathbf{n_k}$ and $\mathbf{n_q}$ are per-pixel normal, and $\mathbf{N}_p$ is the amount of sampled normals.

\noindent\textcolor{black}{\textbf{Mirror Plane Estimation.}}\label{sec:Mirror Plane Estimation}  
Following ~\cite{jin2021planar}, we define the mirror plane $\hat{\pi}=[\mathbf{n}, o]$  using a unit normal vector $\mathbf{n}$ and offset $o$ giving the canonical plane equation $\hat{\mathbf{\pi}}^{T}[x, y, z, -1] = 0 $. 
After optimizing the 3D-GS scene with depth and normals regularization for a few thousand steps, we randomly pick a camera view containing a mirror within its view frustum and backproject the depth $\hat{D}$ and normal $\hat{\mathbf{N}}$ maps of the mirror region to the world coordinate to get 3D mirror coordinates $\hat{C}$ and normal $\mathbf{\hat{N}}_w$. The average value of the normals $\mathbf{\hat{N}}_w$ is considered as the normal of the mirror plane $\mathbf{n}$. We utilize RANSAC~\cite{fischler1981random} and the plane definition to estimate a robust plane equation $\hat{\pi}^{T}[x, y, z, -1] = 0 $. The detailed RANSAC process is presented in the supplementary. Note that we only calculate the mirror plane once during the full training procedure.

\subsection{Virtual Camera Optimization} \label{sec:Virtual Camera Optimization}
We initialize the mirror plane estimate in Sec. \ref{sec:Mirror Plane Estimation}, and use it to compute the virtual camera pose as stated in Sec. \ref{sec:Virtual Camera Rendering}. However, due to the lack of ground truth depth or normal supervision during training, this estimated mirror plane is imperfect. This may lead to a discrepancy between the computed pose of the virtual camera and the ideal pose, resulting in misaligned reflection rendering images. To address this issue, we implemented a simple yet effective method to optimize the estimated mirror plane using virtual camera pose optimization.
Specifically,  in this stage, we optimize the $  \hat{C}_{\text{pred,f}} $ obtained from Sec.\ref{sec:Virtual Camera Rendering} with the photometric loss
between $ \hat{C}_{\text{pred,f}}$ and ground truth image $C_{gt}$:
\begin{equation}\label{eq: vco}
\mathcal{L}_{vco} = \mathcal{L}_{rgb}(\hat{C}_{\text{pred,f}},C_{gt}) .
\end{equation}
During the virtual camera rendering process, we do not compute gradients to the Gaussian attributes ($\mu, \Sigma, O, \theta$). Instead, following iComMa~\cite{sun2023icomma}, we extend 3D-GS by explicitly deriving gradient flow to the virtual camera poses $ \left [ \bf{R}' | \bf{t}' \right ]$. The pose is then optimized during training. Since the virtual camera pose is derived from the mirror plane and the real camera pose, the optimization of the virtual camera pose using the chain rule naturally leads to the refinement of the mirror plane equation as well. This method ensures that both the virtual camera pose and the mirror plane equation are both effectively optimized with the same photometric loss defined above.

We believe that the main bias in virtual camera rendering at this stage comes from the deviation of the virtual camera pose. Additionally, the optimization speed of Gaussians is much faster than that of the camera pose. Consequently, if we optimize the Gaussians' attributes and the virtual camera pose simultaneously, the optimization process prefers duplicating more Gaussians to accommodate the imperfections in the virtual camera pose, rather than correcting the pose itself. This could lead to artifacts in the rendered images. By focusing solely on calculating and optimizing the gradient of the virtual camera pose, we ensure that the optimization process prioritizes the accuracy of the virtual camera pose and the mirror plane equation, thereby enhancing the overall quality of the rendering.

\subsection{Progressive Training}\label{sec:Progressive Training}
We partition our training into various stages. At the beginning of optimization, we train on the full training image dataset with no mirror masking using the original regularization loss proposed in 3D-GS \cite{kerbl20233d}. This ensures that the 3D-GS scene is initialized well. We then activate our depth regularization with Eq. (\ref{eq: depth render}), (\ref{eq: depth smooth}) and normal regularization with Eq. (\ref{eq: loss normal gradient}), (\ref{eq: normal consistent}) still using the full training images. During this stage, we train with the following mirror plane prediction loss:
\begin{equation}\label{eq: l2 loss}
\mathcal{L}_{mpp} = \mathcal{L}_{rgb}(\hat{C}_{\text{pred,r}},C'_{gt}) + \lambda_n\mathcal{L}_{n} + \lambda_s\mathcal{L}_{smooth} + \lambda_{pc}\mathcal{L}_{pc}.
\end{equation}

After this stage, we estimate the mirror plane and enable the virtual camera optimization illustrated in Sec. \ref{sec:Virtual Camera Optimization} and use $\mathcal{L}_{vco}$ in Eq. (\ref{eq: vco}) to optimize the virtual camera pose, hence refine the mirror plane. In this stage, we disable gradients to the Gaussian attributes. In the final stage, we assume that the current mirror plane equation and virtual camera poses have converged to accurate estimates. Therefore, we disable virtual camera pose optimization and optimize the Gaussian scene attributes. During this phase, we fine-tune both non-reflective and reflective regions, leading to photo-realistic novel-view synthesis.

\section{Experiments}

\subsection{Training Details}

Most of the experiments conducted in this paper are on a single V100 GPU and using the pytorch framework (Only the rendering speed (FPS) is tested on a single RTX 4090 GPU), with a total of 60,000 training steps. For all the experiments, the regularization weights are set as follows: $\lambda_{s} = 0.01$, $\lambda_{n} = 0.005$, $\lambda_{pc} = 0.01$, and $\gamma = 0.1$. We employed the Adam Optimizer~\cite{kingma2014adam} for virtual camera optimization, setting the learning rate at 0.0005. The total number of steps for the mirror plane prediction stage was 1000 for experiments conducted on Mirror-NeRF's synthetic dataset and our dataset, and 1500 for Mirror-NeRF's real dataset. The virtual camera optimization stage consisted of 10,000 steps. Please refer to the supplementary for the implementation details.

\subsection{Experimental Results}

\noindent
\textbf{Dataset.} We evaluate our method using the publicly available Mirror-Nerf dataset~\cite{zeng2023mirror}, which includes three synthetic scenes (Washroom, Living room, Office) and three real scenes (Market, Lounge, Discussion room). Each scene is captured within a room containing mirror, with approximately 300 frames taken from a 360-degree perspective. The dataset also includes a mirror reflection mask in each image. 
To enhance the evaluation of our method in real-world scenarios, we provide an additional dataset containing three scenes, Recovery Room, Work Space, and Corridor, with the mirror size and shape in each scene being different from others. The dataset comprises 250 frames per scene, with mirror reflection masks extracted using SAM~\cite{kirillov2023segment} and camera poses estimated by COLMAP~\cite{schonberger2016structure}. We follow ~\cite{zeng2023mirror} to split the training and test dataset and each set includes a mix of views with and without mirror. We show the example images of the mirror scenes from simple to complex in the supplementary.

\noindent
\textbf{Baselines and Evaluation Protocol.} We consider the following methods as baselines for comparison (a) start-of-the-art Mirror-NeRF~\cite{zeng2023mirror}; two fast-rendering methods (b) InstantNGP~\cite{muller2022instant} and  (c) DVGO~\cite{sun2022direct} and (d) the official 3DGS~\cite{kerbl20233d}. Following the standard practice in novel view synthesis, we evaluate the rendering quality using the metrics PSNR, SSIM~\cite{wang2004image} and LPIPS~\cite{zhang2018unreasonable}.

\subsubsection{Mirror-NeRF Dataset Results}
Table.~\ref{tab:mirror-nerf} presents the experimental results for Mirror-Nerf~\cite{zeng2023mirror} synthetic and real datasets. On the synthetic dataset, our method outperforms the best method, Mirror-NeRF, in both PSNR and SSIM metrics, achieving state-of-the-art performance. This success is largely due to our mirror plane prediction described in Section \ref{sec:Mirror Plane Prediction}, which effectively initializes the plane equation on synthetic datasets. The subsequent virtual camera optimization further enhances the rendering quality. Compared to the vanilla 3D-GS, our method achieves significant improvement with the average values across three scenes. Specifically, in the office scene, our method shows gains of +8.65 in PSNR, +0.057 in SSIM, and -0.067 in LPIPS.

\begin{figure*}[t!]
\begin{center}
\includegraphics[width=1.0\linewidth]{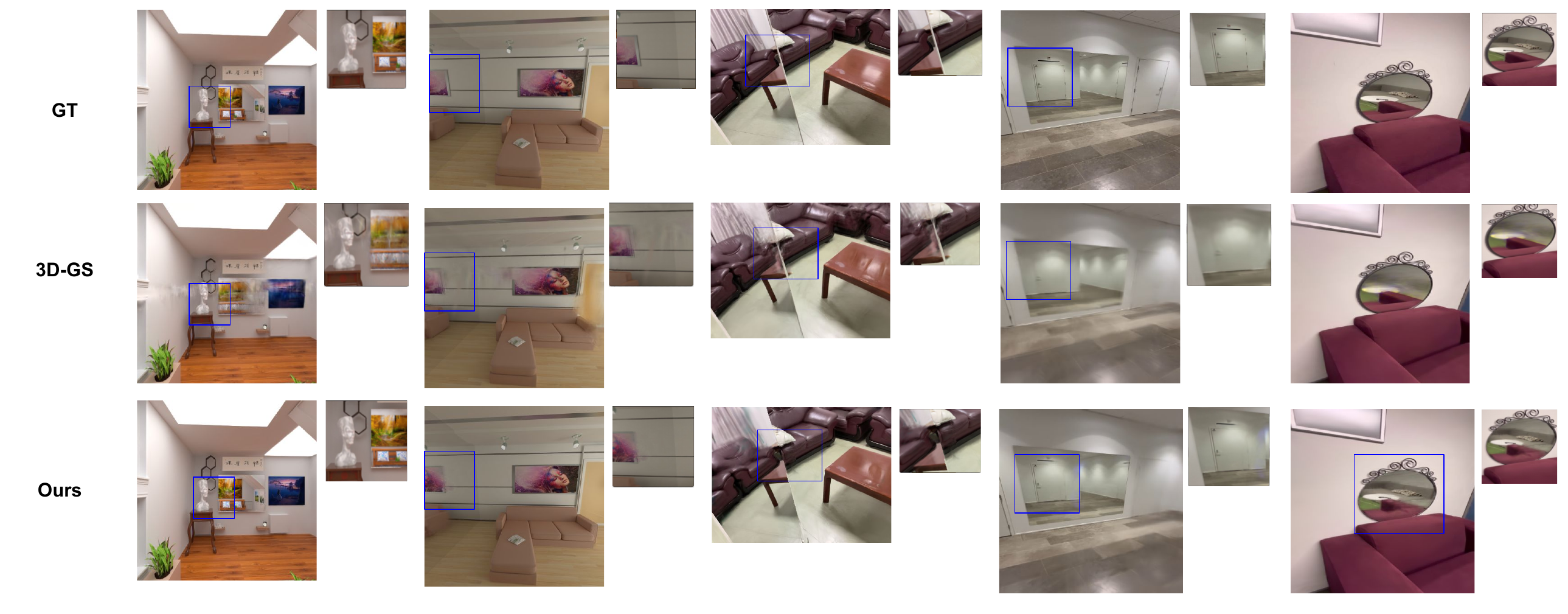}
\vspace{-1em}
\end{center}
   \caption{Qualitative comparisons of ground truth, 3D-GS, and our method on the Living Room, Office, Lounge, Corridor, and Recovery Room scenes from Mirror-NeRF dataset. The smaller image in the upper right corner of each main images is an enlargement of a mirror region.}
\label{fig:vis.pdf}
\end{figure*}

On real datasets,  our method approaches the performance of Mirror-NeRF, with significantly faster rendering speed (FPS 128.22 v.s. 0.71). Compared to more efficient rendering methods like InstantNGP~\cite{muller2022instant} and DVGO~\cite{sun2022direct}, the proposed approach leads in all metrics with a large margin, indicating that our method can provide both high-quality and efficient image rendering. We provide the qualitative comparison in Fig.~\ref{fig:vis.pdf}, and our method achieves high rendering quality with less blurred regions. 

\subsubsection{Our Real-World Dataset Results}

Table.~\ref{tab:custom} presents the comparative results of vanilla 3D-GS and our method on the proposed dataset. In the Corridor and Recovery Room scene which contains large mirrors and planer surface, our method significantly outperforms 3D-GS on all the metrics. 3D-GS estimates mirror reflections as virtual spaces behind the mirror plane, necessitating extensive optimization of additional Gaussians within this virtual space. This results in blurry details and textures, since the resulting mirror regions are not multi-view consistent. Conversely, our method does not require the optimization of additional Gaussians to explain mirror reflections and facilitates more physically accurate mirror rendering using virtual cameras resulting in better detail preservation in the scene.

However, in the Work Space scene, our method exhibits a slight decline relative to 3D-GS. This scene is characterized by a large, cluttered space filled with numerous items, which poses challenges for our method due to the absence of planer surfaces for depth or normal supervision. Consequently, it struggles to derive an effective mirror equation, leading to discrepancies in the virtual camera pose estimation and less effective virtual camera rendering compared to 3D-GS. This highlights areas where further refinement of our method is needed to handle complex scenarios effectively.

\begin{table*}[t!]
\small
\renewcommand\arraystretch{1.3}
\begin{center}
\resizebox{.85\textwidth}{!}
{
\begin{tabular}{lcccclccc}
\toprule 
\multirow{2}{*}{\textbf{Method}} & \multicolumn{4}{c}{\textbf{Synthetic}}             &  & \multicolumn{3}{c}{\textbf{Real}}                 \\
                         & PSNR  $\uparrow$         & SSIM  $\uparrow$         & LPIPS  $\downarrow$        & FPS $\uparrow$ &  & PSNR  $\uparrow$         & SSIM  $\uparrow$         & LPIPS  $\downarrow$           \\ \midrule
InstantNGP~\cite{muller2022instant}               & 23.54          & 0.71           & 0.42           & - &  & 10.51          & 0.20           & 0.71           \\
DVGO~\cite{sun2022direct}                     & 28.05          & 0.82           & 0.29           & - &  & 22.18          & 0.67           & 0.33           \\
Mirror-NeRF~\cite{zeng2023mirror}               & \underline{38.08}    & \underline{0.958}    & \textbf{0.028} & 0.71 &  & \textbf{25.31} & \textbf{0.789} & \textbf{0.082} \\ \midrule
3D-GS~\cite{kerbl20233d}                     & 36.52          & 0.953          & 0.064          & \textbf{480.29} &  & 23.39          & 0.715          & 0.268          \\
\textbf{Ours}            & \textbf{39.87} & \textbf{0.979} & \underline{0.038}    & \underline{128.22} &  & \underline{24.19}    & \underline{0.759}    & \underline{0.234}    \\ \bottomrule

\end{tabular}
}
\end{center}
\caption{Results on the Mirror-NeRF's synthetic and real-world dataset, our method approaches or achieves state-of-the-art performance on these two datasets, and significantly improves upon 3D-GS. The FPS values are measured on an RTX 4090 GPU. The best result is in \textbf{bold}, and the second best is \underline{underlined}.}
\label{tab:mirror-nerf}
\end{table*}

\begin{table}[t!]
\renewcommand\arraystretch{1.3}
\begin{center}
\resizebox{.9\textwidth}{!}
{
\begin{tabular}{lccclccclccc}
\toprule
\multirow{2}{*}{\textbf{Method}}             & \multicolumn{3}{c}{\textbf{Corridor} (easy)}                      &  & \multicolumn{3}{c}{\textbf{Recovery Room} (medium)}        &  & \multicolumn{3}{c}{\textbf{Work Space} (complex)}                           \\
                                     & PSNR $\uparrow$ & SSIM $\uparrow$ & LPIPS $\downarrow$ &  & PSNR $\uparrow$ & SSIM $\uparrow$ & LPIPS $\downarrow$ &  & PSNR $\uparrow$ & SSIM $\uparrow$ & LPIPS $\downarrow$ \\ \midrule
3D-GS~\cite{kerbl20233d} & 25.9           & 0.845           & 0.332              &  & 28.84           & 0.917           & 0.166     &  & \textbf{26.88}  & \textbf{0.884}  & \textbf{0.219}             \\
Ours                                 & \textbf{29.14}  & \textbf{0.874}  & \textbf{0.291}     &  & \textbf{32.21}  & \textbf{0.938}  & \textbf{0.139}              &  & 24.89           & 0.849           & 0.288    \\ \bottomrule

\end{tabular}
}
\end{center}
\caption{Results on the our real-world mirror dataset. Our proposed method outperforms vanilla 3D-GS on two of the scenes. The best result is marked with \textbf{bold}.}
\label{tab:custom}
\end{table}

\subsection{Ablation Study}
We present the ablation study on loss function selection and virtual camera optimization strategies in Table.~\ref{ablation}. The experiments demonstrate that removing any loss function related to the depth map or normal map in plane prediction significantly reduces the accuracy of rendering quality. Removing the normal loss shows the worst performance, indicating the effectiveness of the proposed depth and normal optimization. Additionally, without virtual camera optimization leads to a notable decline in rendering quality. This underscores the critical role of camera optimization in refining both the mirror plane and the virtual camera pose, especially when the initial plane prediction is imperfect. Furthermore, we conduct ablation study on the joint virtual camera and Gaussians optimization. The results demonstrate that this joint optimization detracts from the rendering quality, confirming our assumptions discussed in Sec.~\ref{sec:Virtual Camera Optimization}. 

% In mirror plane estimation, we randomly select a view with a large mirror region and then use this view to estimate the mirror plane. To further investigate whether this random selection affects the quantitative results, we conduct an additional ablation study. Specifically, we utilize the \textit{Living Room} scene from the Mirror-NeRF dataset and randomly choose 5 different views for plane estimation. Each view is trained using the same settings as outlined in our paper, and the metrics are computed as presented in Table~\ref{tab:random}. The results suggest that varying the selected views has minimal impact on the final rendering quality, underscoring the robustness of our approach in accommodating random view selection scenarios.

\begin{table}[t!]
\renewcommand\arraystretch{1.1}
\begin{center}
\resizebox{.6\textwidth}{!}
{
\begin{tabular}{c|ccc}
\toprule
\textbf{Settings}                        & \textbf{PSNR} $\uparrow$          & \textbf{SSIM} $\uparrow$          & \textbf{LPIPS} $\downarrow$         \\ \midrule
w/o $\mathcal{L}_{n}$                     & 32.30          & 0.964          & 0.071          \\
w/o $\mathcal{L}_{pc}$                      & 32.51          & 0.956          & 0.069          \\
w/o $\mathcal{L}_{smooth}$                   & 34.49          & 0.967          & \textbf{0.037} \\
w/o Camera Optimization & 30.79          & 0.943          & 0.102          \\
Camera + Gaussians Optimization & 31.30          & 0.940         & 0.103          \\
Full  Model                     & \textbf{37.30} & \textbf{0.977} & 0.042          \\ 
\bottomrule
\end{tabular}
}
\end{center}
\caption{Ablation studies of various components of our proposed approach using the Office scene form Mirror-NeRF dataset. The highest score is in \textbf{bold}.}
\label{ablation}
\vspace{-10pt}
\end{table}

% \begin{table}[]
% \begin{center}
% \begin{tabular}{cccc}
% \toprule
% \textbf{Scene} & \textbf{PSNR}$\uparrow$ & \textbf{SSIM}$\uparrow$ & \textbf{LPIPS}$\downarrow$ \\ \midrule
% Living\_room   & 43.57 ± 0.321   & 0.992 ± \(6.9 \times 10^{-7}\)   & 0.011 ± \(4 \times 6^{-6}\)    \\ \bottomrule
% \end{tabular}
% \end{center}
% \caption{Ablation study result for different view selection in mirror plane estimation.}
% \label{tab:random}
% \end{table}
\section{Conclusion}
We introduce an improved 3D-GS based method for novel view synthesis in scenes that contain mirror reflections. Our method solves mirror reflections by estimating a mirror plane and modeling reflections using virtual cameras. We extend the 3D-GS representation with depth and normal supervision to accurately estimate the mirror plane in world coordinates. Following this, we employ virtual camera optimization to improve the quality of reflection rendering. To enhance the evaluation of our method, we collect a dataset that includes different mirror sizes in real-world settings. We outperform vanilla 3D-GS and exceed previous NeRF methods on both synthetic and real-world datasets while offering faster rendering speeds. 

\noindent
\textbf{Limitations.} It is important to note that our method assumes the presence of planar surfaces in indoor scenes. The absence of these planar surfaces could lead to a drop in performance due to the inaccurate depth and normal estimates, indicating a need for further methodological enhancements to address more diverse and complex scene geometries. Although our dataset provides more real-world scenes with mirror reflections, it has some limitations. For instance, the placement angles of the mirrors in the scenes are not sufficiently diverse. Future work could consider introducing more challenging real-world datasets to better validate the robustness of the method.

\bibliography{egbib}
\end{document}

\maketitle

%-------------------------------------------------------------------------
\section{Method}
\label{sec:md}

\subsection{Details of plane estimation process}
\label{sec:ransac}
% In Section 2.3, we illustrate how to estimate the mirror plane equation by leveraging the predicted depth map, normal map, and pixel-level mirror region mask. Here is a more detailed explanation of the plane estimation process.

% First, we randomly select a view containing a large mirror region. Then, we obtain the depth and normal pixels in the mirror region using the mask. For the depth map, we project it to points $\hat{C}$ in world coordinates using the camera parameters. For the normal map, since we select the direction of the shortest axis of each Gaussian as the normal when rendering, the rendered normal $\hat{N}_M$ is already in world coordinates. 

% Next, we use the mean of the normals as the normal of the mirror plane, denoted as $\mathbf{n}$. Afterwards, with $\mathbf{n}$ and $\hat{C}$, we use the plane equation to calculate the per-point offset $\hat{O_i}$. To determine the best offset for the mirror plane equation, we implement a RANSAC algorithm on $\hat{O}$ to find the most suitable value for $\mathbf{o}$. This algorithm helps to remove the influence of outliers, ensuring that the selected $\mathbf{o}$ is a robust representative.

In Section 2.3, we illustrate how to estimate the mirror plane equation by leveraging the predicted depth map, normal map, and pixel-level mirror region mask. Here is a more detailed explanation of the plane estimation process.

We first randomly select a view containing a large mirror region and obtain the depth and normal values in the mirror region using the provided mask. Next, the world coordinates $\hat{C}$ are calculated using the estimated depth map and camera parameters. Each 3D coordinate is associated with a normal value from the corresponding pixel on the normal map $\hat{N}_M$. We then compute the mean of the normal values as the normal of the mirror plane, denoted as $\mathbf{n}$. Finally, we fit the plane equation using $\mathbf{n}$ and the 3D points from $\hat{C}$ in a RANSAC~\cite{fischler1981random} loop to obtain the best plane offset $\mathbf{o}$. During our implementation, a large mirror region means the amount of mirror region pixels in that view should be greater than 30000 pixels and the maximum RANSAC iteration is set to 1000 with the distance threshold for offset is 0.1. The detailed process is presented in Algorithm~\ref{alg:plane_fitting}.

\begin{algorithm}[H]
\SetAlgoLined
\KwIn{Depth map $\hat{D}$, Normal map $\hat{N}$, Mirror region mask $M$, extrinsic matrix $ \left [ \bf{R} | \bf{t} \right ]$, intrinsic matrix $\bf{K}$}

\KwOut{Predicted plane equation $\hat{\pi}=[\mathbf{n}, \mathbf{o}]$}
\BlankLine
Extract $\hat{D}_M$, $\hat{N}_M$ from $\hat{D}$, $\hat{N}$ using $M$   \tcp*{$\hat{D}_M$, $\hat{N}_M$ are depth and normal map on mirror region}\ 
$\hat{C} \gets \text{ProjectToWorld}(\hat{D}_M, \mathbf{R}, \mathbf{t}, \mathbf{K})$\ 
\tcp*{$\hat{C}$ is mirror region points in world coordinates}\ 
$\mathbf{n} \gets \text{Mean}(\hat{N}_M)$\ 
\tcp*{Plane Normal Prediction} \
% $\hat{O_i} \gets \mathbf{n}\hat{C_i}$\ 
% \tcp*{Compute per-point offset} \
$\mathbf{o} \gets \text{RANSAC}(\mathbf{n}, \hat{C})$ \tcp*{Plane Offset prediction}
\Return $\hat{\pi}=[\mathbf{n}, \mathbf{o}]$
\caption{Plane estimation}
\label{alg:plane_fitting}
\end{algorithm}

\section{Experiment}
\subsection{Additional implementation details}
% For input data, the batch size is set to 1 for all scenes. Each batch contains a ground truth image, the corresponding mirror region mask, and the camera pose. During training, we generally use the same image size as the input image for most scenes. However, if the image width exceeds 1600 pixels, we resize it to 1600 pixels, as done in 3D-GS~\cite{kerbl20233d}.

% During testing, we resize the rendered images to match the shape used by Mirror-NeRF~\cite{zeng2023mirror} for its synthetic and real datasets. For our dataset, we resize all rendered images to 400 × 400 pixels when computing metrics.

% In mirror plane estimation, we randomly select a view containing large mirror region. During our implementation large mirror region means the amount of mirror region pixels in that view should be greater than 30000 pixels. As for the RANSAC discussed in Sec.\ref{sec:ransac}, we set the iterations = 1000, distance threshold for offset = 0.1. 

% The official implementation of 3D-GS assumes that the $c_x$ and $c_y$ values are always at the center of the image. However, this assumption can lead to blurry images in real scenes captured by phone cameras. To address this issue, we extend the 3D-GS baseline by incorporating $c_x$ and $c_y$ as inputs when calculating the intrinsic matrix for both Mirror-NeRF's real dataset and our dataset.

During training, the batch size is set to 1 for all scenes. Each batch contains an RGB image, the corresponding mirror region mask, and the camera pose. The maximum width of the image is set to 1600px following the protocol in 3D-GS~\cite{kerbl20233d}. During testing, we resize the rendered images to match the shape used by Mirror-NeRF~\cite{zeng2023mirror} for its synthetic and real datasets. In our dataset, we resize all rendered images to 400 × 400 pixels to compute the evaluation metrics. 

The official implementation of 3D-GS assumes that the $c_x$ and $c_y$ values are always at the center of the image. However, this assumption can lead to blurry images in real scenes captured by phone cameras. To address this issue, we extend the 3D-GS baseline by incorporating $c_x$ and $c_y$ as inputs when calculating the intrinsic matrix for both Mirror-NeRF's real dataset and our dataset.

\subsection{Additional details of our dataset}

\begin{figure*}[t!]
\begin{center}
\includegraphics[width=1.0\linewidth]{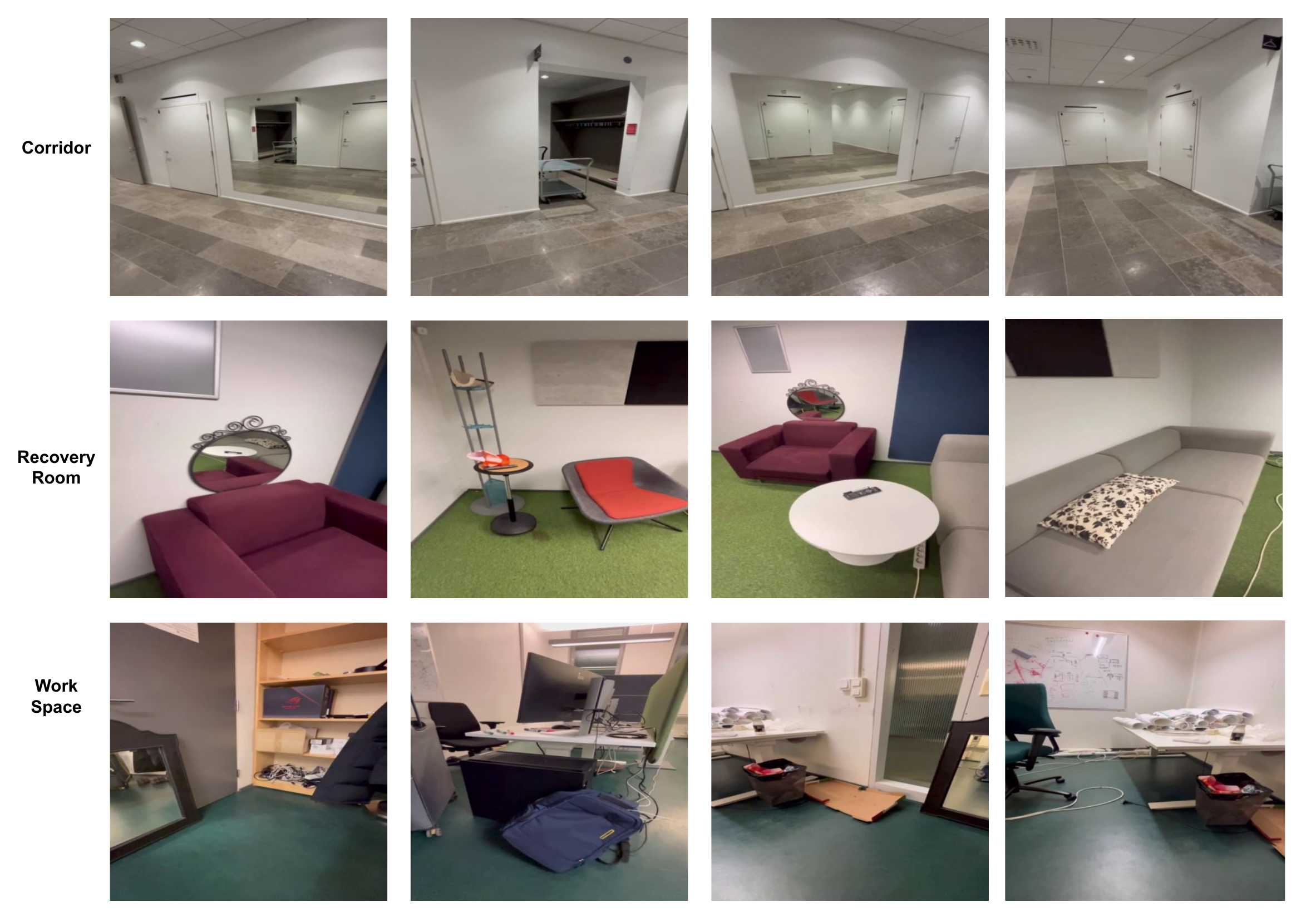}
\end{center}
   \caption{The visualization of our dataset.}
\label{fig:dataset}
\end{figure*}

We propose a dataset that encompasses scenes of different scales and various mirror shapes. \textit{Corridor} features a large mirror on the wall. This scene is relatively straightforward for reflection-based methods as it provides sufficient planar surfaces from the wall, doors, and ground. \textit{Recovery Room} is captured in a small-scale room. In this scene, a small circular mirror is placed on the sofa. It is more complex than the \textit{Corridor}, containing several pieces of furniture such as a sofa, television, and desk. However, most objects have planar or smooth surfaces, which facilitates depth and normal regularization based on planar surface assumptions. \textit{Work Space} includes various irregular objects like cables, clothes, poster rolls, and small items on the desk. A medium-sized mirror is placed along the door. The lack of planar surfaces can result in incomplete 3D reconstruction and challenges in depth and normal estimation, making it difficult to predict and optimize the virtual camera pose. The diversity in scene complexity and mirror shapes in this dataset provides a robust testbed for evaluating reflection-based methods in different environments.

\subsection{Comparison result on our dataset}

\begin{table}[t!]
\renewcommand\arraystretch{1.3}
\begin{center}
\resizebox{.9\textwidth}{!}
{
\begin{tabular}{lccclccclccc}
\toprule
\multirow{}{}{Methods}                                & \multicolumn{3}{c}{Corridor (easy)}                    &  & \multicolumn{3}{c}{Recovery Room (medium)}             &  & \multicolumn{3}{c}{Work Space (complex)}               \\
                                                        & PSNR $\uparrow$ & SSIM $\uparrow$ & LPIPS $\downarrow$ &  & PSNR $\uparrow$ & SSIM $\uparrow$ & LPIPS $\downarrow$ &  & PSNR $\uparrow$ & SSIM $\uparrow$ & LPIPS $\downarrow$ \\ \midrule
3D-GS~\cite{kerbl20233d}          & 25.9            & 0.845           & 0.332              &  & 28.84           & 0.917           & 0.166              &  & \textbf{26.88}  & \textbf{0.884}  & \textbf{0.219}     \\
Mirror-NeRF~\cite{zeng2023mirror}  & 24.61           & 0.821           & 0.396              &  & 25.65           & 0.878           & 0.260              &  & 17.51           & 0.674           & 0.493              \\
Ours                                                    & \textbf{29.14}  & \textbf{0.874}  & \textbf{0.291}     &  & \textbf{32.21}  & \textbf{0.938}  & \textbf{0.139}     &  & 24.89           & 0.849           & 0.288              \\ \bottomrule
\end{tabular}
}
\end{center}
\caption{Results on the our real dataset, Our method outperforms 3D-GS on two of the scenes and outperforms Mirror-NeRF in all three scenes. The best result is in \textbf{bold}.}
\label{tab:custom}
\end{table}

We have extended the quantitative results for our dataset by including the performance of Mirror-NeRF~\cite{zeng2023mirror}. Our findings indicate that Mirror-NeRF struggles to accurately predict the normal and depth of mirror regions in our proposed dataset, resulting in blurry renderings of mirror region. As shown in Table~\ref{tab:custom}, our method outperforms Mirror-NeRF across all three scenes.

\begin{figure*}[t!]
\begin{center}
\includegraphics[width=1.0\linewidth]{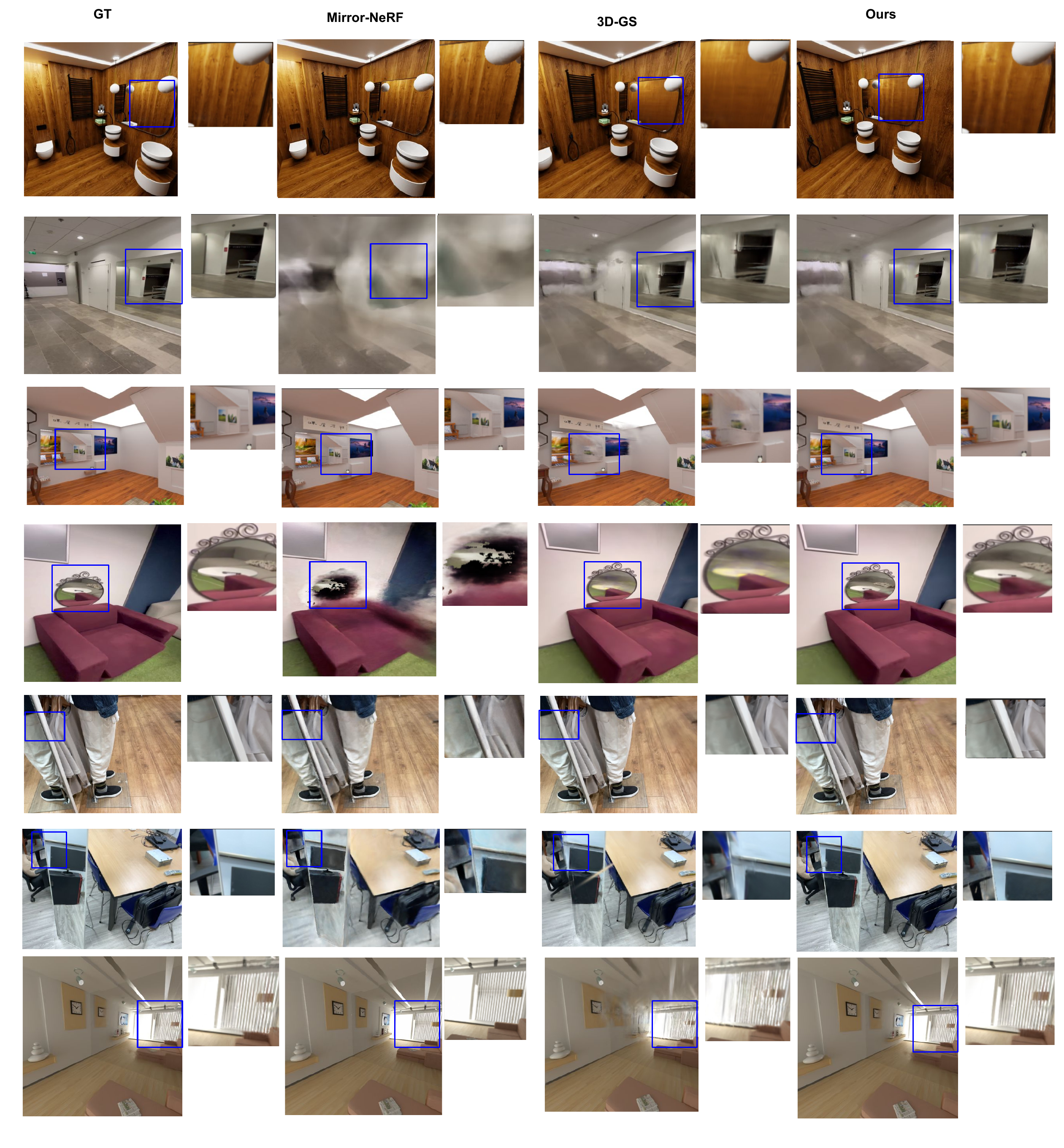}
\end{center}
   \caption{The visual comparison of GT, Mirror-NeRF, 3D-GS, and ours method. The smaller image in the upper right corner of each main images is an enlargement of a mirror region.}
\label{fig:vis_sup.pdf}
\end{figure*}

Additional visualization results on both Mirror-NeRF's dataset and our dataset are presented in Fig.~\ref{fig:vis_sup.pdf}.

\subsection{Ablation study for mirror plane estimation}
In mirror plane estimation, as described in the main paper, we randomly select a view with a large mirror region and then use this view to estimate the mirror plane. To further investigate whether this random selection affects the quantitative results, we conduct an additional ablation study. Specifically, we utilize the \textit{Living Room} scene from the Mirror-NeRF dataset and randomly choose 5 different views for plane estimation. Each view is trained using the same settings as outlined in our paper, and the metrics are computed as presented in Table~\ref{tab:random}. The results suggest that varying the selected views has minimal impact on the final rendering quality, underscoring the robustness of our approach in accommodating random view selection scenarios.

\begin{table}[]
\begin{center}
\begin{tabular}{cccc}
\toprule
Scene & PSNR$\uparrow$ & SSIM$\uparrow$ & LPIPS$\downarrow$ \\ \midrule
Living\_room   & 43.57 ± 0.321   & 0.992 ± \(6.9 \times 10^{-7}\)   & 0.011 ± \(4 \times 6^{-6}\)    \\ \bottomrule
\end{tabular}
\end{center}
\caption{Ablation study result for different view selection in mirror plane estimation.}
\label{tab:random}
\end{table}

%-------------------------------------------------------------------------

%------------------------------------------------------------------------

\bibliography{egbib}